\documentclass[10pt,twocolumn,letterpaper]{article}

\usepackage{cvpr}              % To produce the CAMERA-READY version
\usepackage{amsmath,amsfonts,bm}

\def\eqref#1{equation~\ref{#1}}
\def\1{\bm{1}}

\DeclareMathAlphabet{\mathsfit}{\encodingdefault}{\sfdefault}{m}{sl}
\SetMathAlphabet{\mathsfit}{bold}{\encodingdefault}{\sfdefault}{bx}{n}

\usepackage{graphicx,adjustbox}
\usepackage{url}
\usepackage{amssymb}
\usepackage[linesnumbered,ruled,vlined]{algorithm2e}
\SetAlFnt{\small}
\SetKwComment{Comment}{/* }{ */}
\SetKwInput{kwInput}{Input}
\SetKwInput{kwOutput}{Output}
\RestyleAlgo{ruled}
\usepackage{amsmath}
\usepackage{array}
\usepackage{multirow}
\usepackage{booktabs}
\usepackage[table]{xcolor}
\usepackage[normalem]{ulem}
\useunder{\uline}{\ul}{}
\usepackage{float, wrapfig}
\usepackage{resizegather}
\usepackage{enumitem}

\usepackage{cuted}
\usepackage{capt-of}

\usepackage[accsupp]{axessibility}

\usepackage[font=small,skip=0pt]{caption}
\definecolor{cvprblue}{rgb}{0.21,0.49,0.74}
\usepackage[pagebackref,breaklinks,colorlinks,allcolors=cvprblue]{hyperref}

\def\paperID{35812} % *** Enter the Paper ID here
\def\confName{CVPR}
\def\confYear{2026}

\title{Test-Time Instance-Specific Parameter Composition: A New Paradigm for Adaptive Generative Modeling}

\author{Minh-Tuan Tran$^1$, Xuan-May Le$^2$, Quan Hung Tran$^3$, Mehrtash Harandi$^1$, Dinh Phung$^1$, Trung Le$^1$\\\
$^1$Monash University,  $^2$The University of Melbourne, $^3$Meta Inc\\
{\tt\small \{tuan.tran7,mehrtash.harandi,dinh.phung,trunglm\}@monash.edu} \\ \tt\small xuanmay.le@student.unimelb.edu.au, quanhungtran@meta.com}

\begin{document}
\maketitle
\begin{abstract}
Existing generative models, such as diffusion and auto-regressive networks, are inherently static, relying on a fixed set of pretrained parameters to handle all inputs. In contrast, humans flexibly adapt their internal generative representations to each perceptual or imaginative context. Inspired by this capability, we introduce Composer, a new paradigm for adaptive generative modeling based on test-time instance-specific parameter composition. Composer generates input-conditioned parameter adaptations at inference time, which are injected into the pretrained model’s weights, enabling per-input specialization without fine-tuning or retraining. Adaptation occurs once prior to multi-step generation, yielding higher-quality, context-aware outputs with minimal computational and memory overhead. Experiments show that Composer substantially improves performance across diverse generative models and use cases, including lightweight/quantized models and test-time scaling. By leveraging input-aware parameter composition, Composer establishes a new paradigm for designing generative models that dynamically adapt to each input, moving beyond static parameterization. The code will be available at \url{https://github.com/tmtuan1307/Composer}.
\end{abstract}

\section{Introduction}

Generative models such as diffusion \cite{ldm, dit, controlnet} and visual auto-regressive models (AR) \cite{var, LlamaGen, infinity, infinitystar} have achieved remarkable progress in synthesizing high-fidelity and diverse images. Despite this success, most existing models remain static, a single, fixed set of parameters must accommodate all prompts, scenes, and modalities. This rigidity fundamentally limits adaptability: while humans dynamically adjust their internal generative representations to each perceptual or imaginative context \cite{composer-moti1,composer-moti2}, current generative models rely on immutable weights that cannot specialize to the nuances of each input. As a result, they often produce oversmoothed or inconsistent samples under complex or ambiguous conditions.

Recent advances in test-time training~\cite{ttt1,ttt2,ttt3, ttt4} demonstrate that adapting model parameters at inference time can improve performance. While such techniques have become increasingly popular in large language models, they remain largely unexplored for image generative models, where high-resolution diffusion and VAR backbones already demand massive computational budgets, making instance-wise gradient optimization at test time prohibitively expensive. Mixture-of-Experts (MoE) architectures~\cite{diff-moe1, diff-moe2, diff-moe3, llm-moe1} provide conditional computation by activating different experts per input, but their routing is coarse-grained and tied to a fixed expert pool, limiting instance-specific adaptation. In addition, they usually entail architectural changes and full retraining, rather than being applicable to off-the-shelf pretrained models.

\begin{figure}[t]
\begin{center} 
\includegraphics[width=\linewidth]{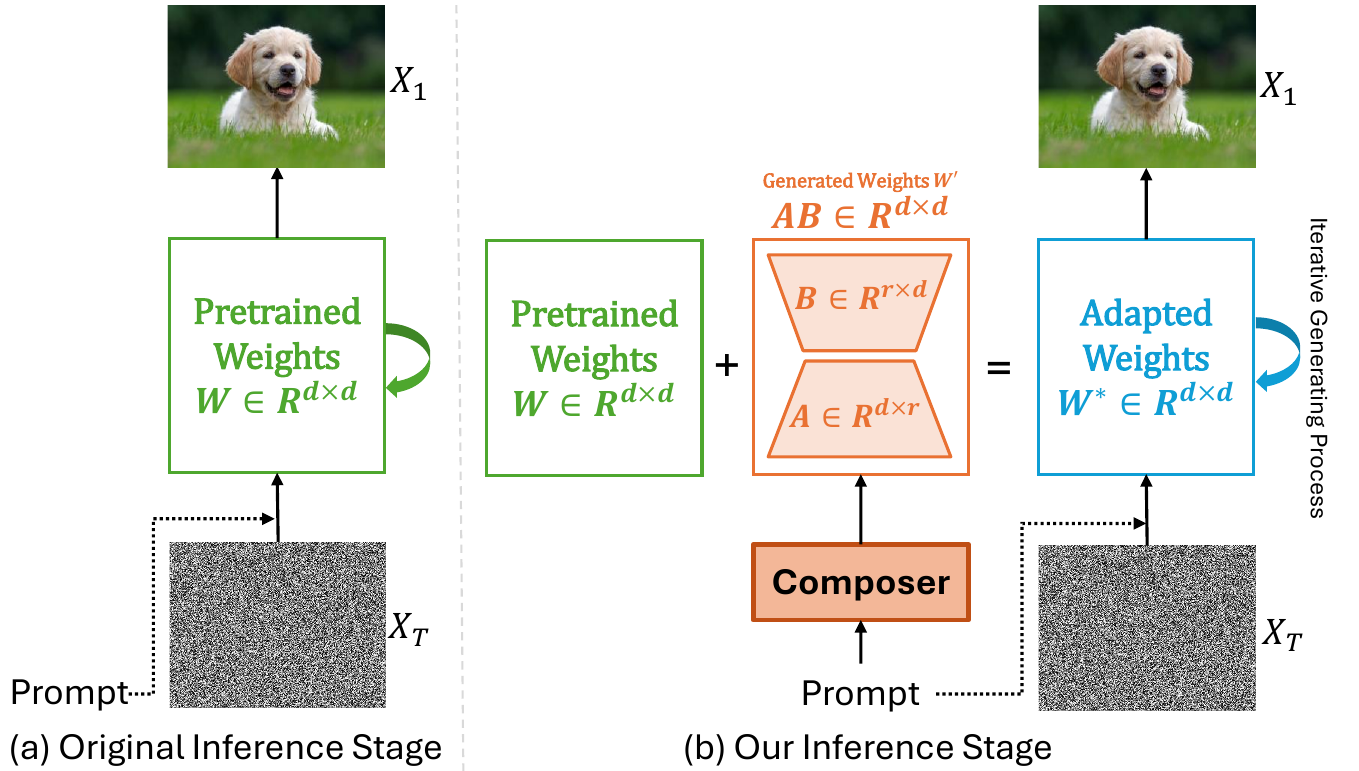}
\end{center}
\caption{Comparison of static versus adaptive parameterization. Composer dynamically composes instance-specific parameter updates, allowing per-input adaptation without fine-tuning.}
\label{fig:nrrdd}
\end{figure}

\begin{table}[t]
\centering
\caption{Comparison of average generation quality, inference time, and peak memory for class-conditional image generation on ImageNet $256\times256$. It is clear that Composer improves FID while introducing negligible overhead in inference time and memory, unlike test-time training which is significantly more expensive.}

\adjustbox{max width=\linewidth}{
\begin{tabular}{lccc}
\toprule
Method & FID $\downarrow$ & Inference Time (s) & Peak Memory (GB) \\
\midrule
Standard & 3.03 & 15.6 & 7.33 \\
Test-time Training & 2.86 & 84.24 (+540\%) & 13.26 (+180\%) \\
Composer & \textbf{2.77} & \textbf{15.63 (+0.2\%)} & \textbf{7.59 (+3.6\%)} \\
\bottomrule
\end{tabular}
}
\label{tab:fid_memory_time}
\end{table}

In this work, we move beyond these limitations by endowing generative models with the ability to compose parameters dynamically for each input. Instead of iterative fine-tuning, we propose to synthesize instance-specific parameter adaptations directly from the conditioning signal through a lightweight auxiliary network. These low-rank updates are composed with the pretrained model weights, forming an adaptive configuration that better aligns with the semantics of each input.

We introduce Composer, a plug-in meta-generator that produces compact, input-conditioned parameter updates for arbitrary pretrained generators at inference time. The generated updates serve as adaptive bridges between the conditioning features and the pretrained parameters, enabling per-input specialization without additional optimization. Composer performs this parameter composition once prior to generation, introducing negligible computational overhead while significantly improving fidelity and consistency.

Our framework is model-agnostic and integrates seamlessly into various generative backbones, including diffusion and visual auto-regressive models. Empirically, Composer enhances generation quality, controllability, and context alignment across diverse settings. Furthermore, its adaptive mechanism generalizes to lightweight and quantized backbones, partially restoring generation quality while maintaining efficiency.

\textbf{Our main contributions are as follows:}
\begin{itemize}
    \setlength\itemsep{0pt} % remove space between items
    \setlength\parskip{0pt} % remove paragraph spacing
    \item We propose \textbf{Composer}, a plug-in framework for test-time instance-specific parameter composition, enabling pretrained generative models to adapt their weights dynamically per input at inference \emph{without backbone's architectural changes or exhaustive retraining}.
    \item We introduce a meta-generator transformer that maps input conditions to low-rank parameter updates, enabling efficient, context-aware adaptation while leveraging frozen pretrained weights. Moreover, we design a novel training pipeline that encourages consistent adaptation for semantically similar inputs and preserves contrast across different contexts, improving stability and generalization.
    \item We demonstrate that Composer consistently improves the performance of diverse generative backbones across class-conditional image generation, text-to-image synthesis, post-training quantization, and test-time scaling for diffusion models.
\end{itemize}

By enabling input-aware parameter composition, Composer advances the field toward adaptive generative modeling, where models can reconfigure their internal parameters on-the-fly, akin to how human cognition dynamically adapts to context and imagination.

\section{Adaptive Generative Modeling with Composer}

\begin{figure}[t]
\begin{center}
\includegraphics[width=\linewidth]{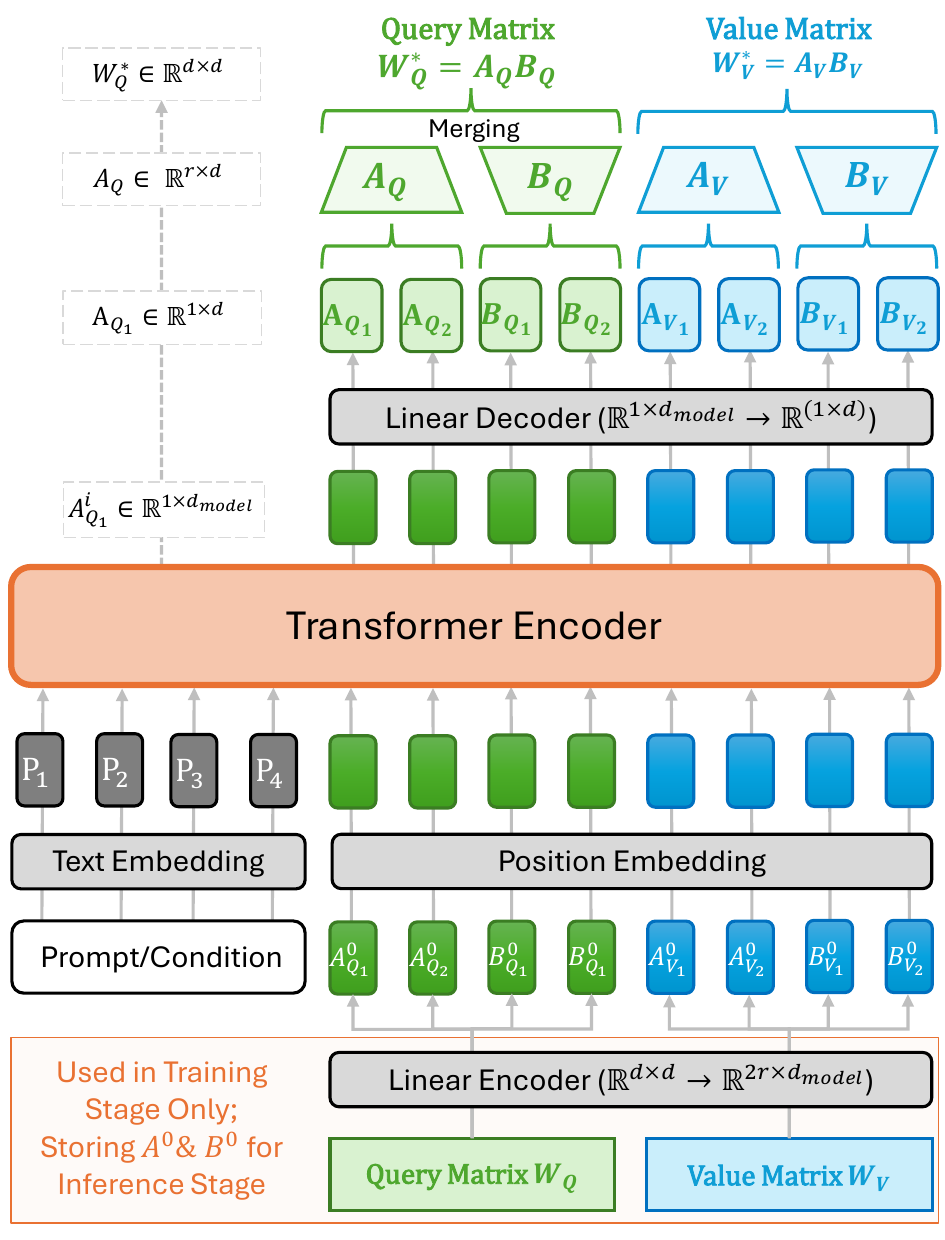}
\end{center} 
\caption{
\textbf{Overview of Composer.} Given any weight matrix $W$ from the backbone, Composer generates a low-rank update $W' = W +  AB$ conditioned on the input. Specifically, the query and value matrices $W_Q$ and $W_V$ from the pretrained model are linearly projected from $\mathbb{R}^{d \times d}$ to $\mathbb{R}^{2r \times d_{\text{model}}}$. The projected representations are then separated to initialize tokens $A^0_i$ and $B^0_i \in \mathbb{R}^{1 \times d_{\text{model}}}$. During training, these tokens are combined with prompt tokens $P_i \in \mathbb{R}^{1 \times d_{\text{model}}}$ and processed by a transformer to produce $W^* = AB$. The adapted parameters $W' = W + W^*$ are used for generation. At inference, the first projection layers are removed, while $A^0_i$ and $B^0_i$ are stored for fast instance-specific adaptation. 
}
\label{fig:composer}

\end{figure}

Existing generative models, such as diffusion and auto-regressive networks, are inherently \textit{static}, relying on a fixed set of pretrained parameters to handle all inputs. In contrast, humans flexibly adapt their internal representations for each perceptual or imaginative context. Inspired by this capability, we propose \textbf{Composer}, a new paradigm for \textit{adaptive generative modeling} based on \textit{instance-specific parameter composition}. 

Composer generates input-conditioned parameter adaptations at inference time, which are injected into the pretrained backbone weights, enabling per-input specialization without fine-tuning or retraining. Adaptation occurs once prior to multi-step generation, producing higher-fidelity, context-aware outputs with minimal computational and memory overhead. Figure~\ref{fig:nrrdd} illustrates the Composer workflow, showing how instance-specific updates $W^*$ are generated, composed with the pretrained weights $W$, and used in diffusion-based generation.

\subsection{Instance-Specific Parameter Composition in Generative Models}

Modern generative architectures, such as diffusion and auto-regressive networks, consist of deep backbones parameterized by numerous weight matrices $W$. 
In diffusion models, $W$ typically represents the linear projections for the attention components (i.e., $W_Q$, $W_V$, $W_V$) or the convolutional filters within the U-Net. In this work, we primarily apply our method to the query ($W_Q$) and value ($W_V$) projection matrices within transformer-based architectures \cite{transformer, shapeformer}, as these components have proven most effective in prior fine-tuning approaches~\cite{lora1, lora2}. We also conducted an ablation study comparing the effectiveness and efficiency of applying adapters to different combinations of $W_Q, W_K, W_V,$ and $W_O$, as detailed in the \textbf{Supplementary Material}.

% In auto-regressive transformers, $W$ corresponds to the weights of self-attention or feed-forward layers that control token interactions and contextual reasoning.  

Despite their expressiveness, these weights are \textit{frozen} after pretraining: \textit{the same parameters are applied to every input, regardless of content or modality}. 
Such static usage limits contextual adaptability and often requires expensive fine-tuning or specialized adapters to achieve task-specific performance. Moreover, in LoRA \cite{lora1, lora2}, the weight update is defined as $W' = W + AB$, where $A$ and $B$ are two low-rank matrices shared across all input data, hence limiting the ability of $A$ and $B$ to adapt to complex tasks. 

To overcome this limitation, Composer reformulates the weight matrix $W$ as a \textit{composable structure} that can be dynamically adapted for each input instance. 
We introduce two low-rank matrices $A$ and $B$ that define an instance-specific residual update:
\begin{equation}
W' = W + AB,
\end{equation}
where $A \in \mathbb{R}^{d \times r}$, $B \in \mathbb{R}^{r \times d}$, and $r \ll d$. 
Here, $A$ and $B$ are two low-rank matrices generated by inputting the original weight matrix $W$ and an individual input (e.g., a text prompt) into our Composer, serving as lightweight, learnable corrections that modulate the pretrained weights without altering their underlying knowledge. This formulation allows flexible, per-input specialization during inference, maintaining efficiency while enhancing generative fidelity.  

In the following subsection, we describe how Composer leverages a transformer-based generator to produce these instance-specific low-rank matrices conditioned on contextual signals such as prompts or class embeddings.

\subsection{Transformer-Based Parameter Generation}

Composer generates instance-specific parameter updates $W'$ using a low-rank decomposition:
\begin{equation}
W' = W+ AB, \quad A \in \mathbb{R}^{d \times r}, \quad B \in \mathbb{R}^{r \times d},
\end{equation}
where $d$ denotes the dimension of the backbone weight matrix, and $r$ denotes the rank of the update ($r \ll d$). Indeed, we view each $A$ and $B$ as a sequence of $r$ tokens, where each token has a dimension of $d \times 1$.  

To enable fast inference while capturing correlations across weight matrices, we design the instance-specific parameters using a transformer-based model. The architecture of our parameter generator is illustrated in Figure~\ref{fig:composer}.

% This decomposition efficiently captures instance-specific variations while keeping both memory and computation costs low. 

\noindent\textbf{Token Initialization.} 
During training, each token of $A$ and $B$ is initialized from the pretrained weight matrices of the pretrained transformer (e.g., $W_Q$ and $W_V$ with a dimension of $d \times d$) via a linear projection:
\begin{align}
[A^0_{Q_1}, \dots, A^0_{Q_r}, B^0_{Q_1}, \dots, B^0_{Q_r}] &= \text{Linear}(W_Q), \\
[A^0_{V_1}, \dots, A^0_{V_r}, B^0_{V_1}, \dots, B^0_{V_r}] &= \text{Linear}(W_V)
\end{align}
where $\text{Linear}(\cdot)$ maps $\mathbb{R}^{d \times d}$ to $\mathbb{R}^{2r \times d_{\text{model}}}$, producing $r$ tokens for $A$ and $r$ for $B$ in $\mathbb{R}^{1 \times d_{\text{model}}}$. 
This initialization leverages pretrained knowledge, providing a strong starting point for learning. 
Each resulting vector forms a token:
\begin{align}
A^0_Q = \{A^0_{Q_1}, \dots, A^0_{Q_r}\}, \quad 
B^0_Q = \{B^0_{Q_1}, \dots, B^0_{Q_r}\}, \nonumber \\
A^0_V = \{A^0_{V_1}, \dots, A^0_{V_r}\}, \quad 
B^0_V = \{B^0_{V_1}, \dots, B^0_{V_r}\}.
\end{align}
During inference, the linear projection is removed, and the learned $A^0_{Q|V}$ and $B^0_{Q|V}$ tokens are stored for efficient reuse.

\noindent\textbf{Positional Embedding and Prompt Conditioning.} 
Learned positional embeddings are added to each token:
\begin{align}
A^1_{Q_i} = A^0_{Q_i} + PE(A^0_{Q_i}), \; 
B^1_{Q_i} = B^0_{Q_i} + PE(B^0_{Q_i}), \nonumber\\
A^1_{V_i} = A^0_{V_i} + PE(A^0_{V_i}), \; 
B^1_{V_i} = B^0_{V_i} + PE(B^0_{V_i})
\end{align}
and concatenated with \textit{input prompt tokens} $P = [P_1, \dots, P_m]$ during training:
\begin{equation}
X = [P_1, \dots, P_m, A^1_{Q_{1:r}}, B^1_{Q_{1:r}}, A^1_{V_{1:r}}, B^1_{V_{1:r}}].
\end{equation}
Including prompt tokens allows the transformer to condition parameter generation on input context such as textual or semantic cues. 

% During inference, prompt tokens are omitted, and stored $A^*$ and $B^*$ tokens are directly used.
\begin{figure}[t]
\begin{center}
\includegraphics[width=0.8\linewidth]{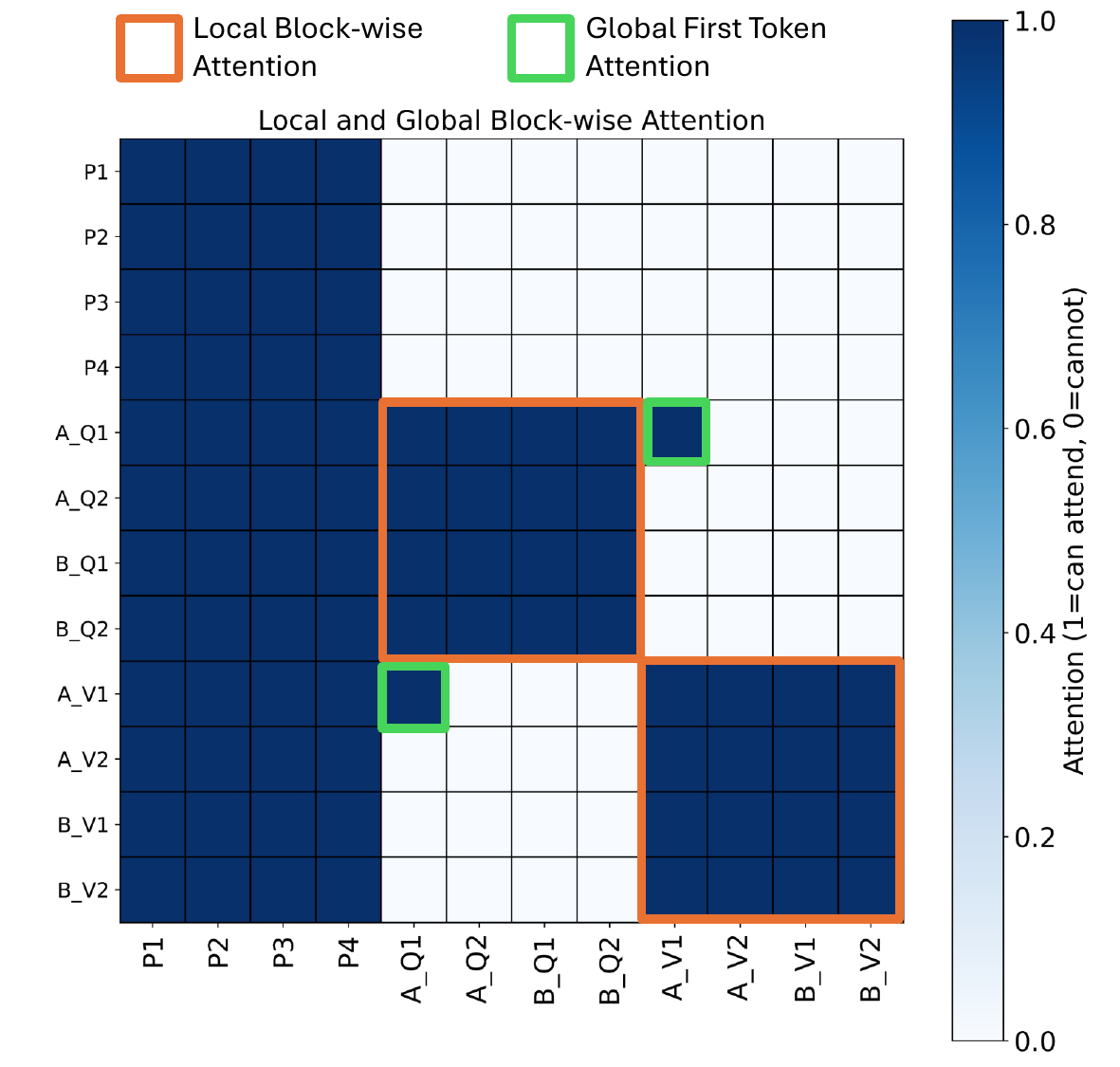}
\end{center} 
    \caption{Illustration of the attention scheme. Component tokens attend to prompt tokens for context, maintain local block-wise attention, and the first token of each block captures inter-block correlations.}
\label{fig:attention}
\end{figure}

\noindent\textbf{Transformer Architecture.} 
Composer employs an encoder-style transformer with multi-head attention and feed-forward layers. 
Attention operates according to the following scheme, as illustrated in Figure~\ref{fig:attention}:
\begin{itemize}
    \setlength\itemsep{0pt} % remove space between items
    \setlength\parskip{0pt} % remove paragraph spacing
    \item All component tokens attend to every prompt token to ensure context-aware adaptation.
    \item Local block-wise attention among component tokens reduces computational cost while preserving intra-block coherence.
    \item The first token from each block attends across blocks to capture inter-block correlations.
\end{itemize}

After processing, the output embeddings at the $L$-th layer corresponding to $A^L_{Q}, B^L_{Q}$ and $A^L_{V}, B^L_{V}$ tokens are reshaped to reconstruct the low-rank matrices $A_Q,B_Q$ and $A_V, B_V$, yielding the instance-specific update:
\begin{equation}
W'_{Q} = W_Q + A_{Q}B_{Q} \text{  and  } W'_{V} = W_V + A_{V}B_{V}. 
\end{equation}

\subsection{Weight Composition: Training vs. Inference}

\noindent\textbf{Training Stage.} 
During training, Composer generates instance-specific low-rank matrices $A$ and $B$ conditioned on the prompt $P$:
\begin{equation}
A, B = \text{Composer}(A^0, B^0, P),
\end{equation}
which define the update $W' = W+ AB$. 
The model output is computed as:
\begin{equation}
h = W x + ABx,
\end{equation}
allowing gradients to propagate through $A$ and $B$ while preserving the original pretrained behavior of $W$. 
Prompt tokens and positional embeddings provide contextual conditioning during training.

\noindent\textbf{Inference Stage.} 
At inference, the learned tokens $A^0$ and $B^0$ are used to generate the low-rank updates, which are merged with the pretrained weight:
\begin{align}
A, B &= \text{Composer}(A^0, B^0, P), \\
W' &= W + AB, \;
h = W' x,
\end{align}
enabling efficient, per-input, context-aware generation in a single forward pass.

\subsection{Context-Aware Training Pipeline for Composer}

To train Composer effectively, we introduce a context-aware training strategy that balances consistency and diversity in instance-specific adaptations. We organize this discussion into four parts.

\noindent\textbf{i. Vanilla Training.} 
Standard approaches for training generative models randomly collect $(\text{image}, \text{text})$ or $(\text{image}, \text{class})$ pairs from the dataset. 
While straightforward, this random sampling does not explicitly enforce semantic relationships, which can lead to inconsistent generation across similar inputs and unstable adaptations in instance-specific weights.

\noindent\textbf{ii. Full-Class or Similar-Prompt Sampling.} 
An improvement over random sampling is to select a full set of images from a single class or all samples with semantically similar prompts for each batch. 
This strategy encourages consistent adaptations for inputs sharing the same label or prompt context. 
However, focusing solely on one class or prompt type can reduce diversity and increase the risk of mode collapse across the dataset.

\noindent\textbf{iii. Context-Aware Pipeline for Class-Conditioned Image Generation.} 
Our approach combines the benefits of consistency and diversity by splitting each batch based on a parameter $\alpha \in [0,1]$: 
\begin{itemize}
    \setlength\itemsep{0pt} % remove space between items
    \setlength\parskip{0pt} % remove paragraph spacing
    \item $\alpha \times b$ images are sampled from the same class or with semantically close prompts to enforce consistent instance-specific adaptations.
    \item $(1-\alpha) \times b$ images are sampled from different classes or distant contexts to ensure outputs remain distinct.
\end{itemize}
For a batch of images $x$ with conditioning signals $P$, the adapted weights are generated as:
\begin{equation}
W' = W + AB, \quad A,B = \text{Composer}(A^0, B^0,P).
\end{equation}
The model is trained with the standard diffusion objective:
\begin{equation}
\mathcal{L}_{\text{diffusion}} = \mathbb{E}_{x, \epsilon, t} \left[ \| \epsilon - \epsilon_\theta(x_t, t; W', P) \|_2^2 \right],
\end{equation}
where $x_t = \sqrt{\gamma_t} x + \sqrt{1-\gamma_t} \epsilon$ and $\epsilon \sim \mathcal{N}(0,I)$.  
This structured pipeline balances consistency among related inputs with diversity across contrasting contexts, and outperforms vanilla random-pair training.

\noindent\textbf{iv. Context-Aware Pipeline for Text-to-Image Generation.} 
For text-to-image tasks, we further refine batch selection by measuring semantic similarity in a learned embedding space (e.g., CLIP embeddings).  
\begin{itemize}
    \setlength\itemsep{0pt} % remove space between items
    \setlength\parskip{0pt} % remove paragraph spacing
    \item For each input, we select the top $\alpha \times b$ most similar images in embedding space to encourage consistent adaptations.  
    \item The remaining $(1-\alpha) \times b$ samples are drawn from distant embeddings to enforce diversity.  
\end{itemize}
This dynamic similarity sampling captures semantic relationships beyond exact class labels or prompt overlap, improving the contextual coherence of instance-specific adaptations while preserving output diversity.  
\section{Further Adaptation}

\subsection{Composer for Post-Training Quantization}

Efficient machine learning \cite{kd-survey,nayer,lander,muse,nrrdd,q-diffusion,dmq,} plays a key role in enabling modern deep models to be deployed under limited memory and computation budgets. As a further application of our proposed method, we Composer can be extended to \textit{post-training quantization} (PTQ) of generative models \cite{q-diffusion,q-dit,quest,ctec,vq4dit,dmq}. In quantized models, both weights and activations are reduced to low-precision formats (e.g., INT8 or INT4), which can introduce significant approximation errors and degrade generative fidelity. To address this, we propose a \textit{quantization-aware Composer}, where the low-rank updates and the Composer weights themselves are trained in the same low-precision format as the backbone.

For a quantized backbone weight $W_q$ and full-precision activations $x$, Composer generates instance-specific low-rank updates $A$ and $B$ and a learnable activation scaling factor $\gamma$:
\begin{align}
W'_q &= W_q + AB, \\
x'_q &= \text{Quantize}(\gamma \cdot x), \\
h_q &= W'_qx'_q,
\end{align}
where $x'_q$ is the scaled and quantized activation, and $h_q$ is the resulting hidden representation. The updates $(A, B, \gamma)$ are generated from the stored tokens $(A^0, B^0, \gamma^0)$ conditioned on the input context $P$:
\begin{equation}
[A, B, \gamma] = \text{Composer}(A^0, B^0, \gamma^0, P).
\end{equation}

Training Composer in a quantization-aware manner ensures that the low-rank updates are effective under the low-precision setting, enabling input-dependent compensation for quantization errors. We adopt a knowledge distillation loss from the full-precision teacher model:
\begin{equation}
\mathcal{L}_{\text{KD}} = \| h - h_q \|_2^2,
\end{equation}
where $h = W x$ is the teacher output and $h_q$ is the output of the quantized backbone with Composer updates.

This framework allows Composer to restore the fidelity lost due to both weight and activation quantization, while keeping memory and computation overhead low, and requiring no retraining of the original quantized model. By training Composer in the same low-precision format as the backbone, we ensure that it is fully compatible and effective in practical quantized deployment scenarios.

\subsection{Composer for Test-Time Scaling in Diffusion Models}

Another practical application of Composer is \textit{test-time scaling} for diffusion-based generative models \cite{image-tts,video-tts,var,t2i-tts}. In this setting, the pretrained backbone weights remain fixed, and Composer is used to generate input-conditioned low-rank updates at inference time, allowing adaptive scaling of the model’s behavior for each instance.

In contrast to quantization-aware training, here Composer is applied \textit{post hoc} at test time, without retraining the backbone or the Composer weights. This enables instance-specific adaptation and scaling during inference, providing improved flexibility and higher-fidelity outputs in multi-step diffusion generation.

Moreover, Composer can further enhance test-time scaling by producing refined, input-dependent parameter updates, effectively compensating for limitations of naive scaling strategies. By leveraging these adaptive updates, the model can dynamically adjust its internal representations for each input, improving output quality and main

\section{Experiments}

We evaluate Composer across multiple generative modeling scenarios to demonstrate its effectiveness in instance-specific adaptation, computational efficiency, and high-fidelity generation. Our experiments cover both \textit{class-conditioned} and \textit{text-to-image generation}, as well as practical applications in \textit{quantization-aware} and \textit{test-time scaling settings}.

\noindent\textbf{Experimental Setup:} We compare our method with standard static generative models. To further demonstrate the benefits of Composer, we also conduct experiments using standard test-time training for generative models, where similar data to the input is randomly selected and the model is fine-tuned (using the same low-rank dimension $r$ as in our method) before generating images. In all experiments, we set the low-rank dimension to $r = 8$ and the context-aware sampling ratio to $\alpha = 0.75$.
All experiments are trained using NVIDIA A100 80GB. The training process employs the
AdamW optimizer with a weight decay rate of 0.05. The learning rate is set to 1e-4, with the training epochs are 50 for all experiments. For evaluation, FID and IS (Inception Score) are employed for class-conditional image generation, whereas GenEval is used for text-to-image generation. 

\subsection{Class-Conditional Image Generation}

\begin{table*}[t]
\centering
\caption{Comparison of different generative models and methods in class-conditional image generation on ImageNet $256\times256$. Inference time shows absolute and percentage increase relative to Standard.}
\adjustbox{max width=0.7\textwidth}{
\begin{tabular}{@{}llrrrrrrrr@{}}
\toprule
                           &                    & \multicolumn{1}{c}{FID} & \multicolumn{1}{c}{IS} & \multicolumn{1}{c}{Pre} & \multicolumn{1}{c}{Rec} & \multicolumn{1}{c}{Step} & \multicolumn{1}{c}{Parameter} & \multicolumn{1}{c}{Time} & \multicolumn{1}{c}{Memory} \\ \midrule
                           & Standard           & 3.55 & 274.4 & 0.84 & 0.51 & 10 & 310M & 0.4 & 2.37G \\
                           & Test-time Training \cite{ttt1}& 3.22 & 277.2 & 0.84 & 0.53 & 10 & 310M & 40.52 (+10,030\%) & 4.58G \\
\multirow{-3}{*}{VAR d-16 \cite{var}} & Composer           & \textbf{3.15} & \textbf{280.4} & \textbf{0.85} & \textbf{0.53} & \textbf{10} & \textbf{412M} & \textbf{0.42 (+5\%)} & \textbf{2.57G} \\ \midrule
                           & Standard           & 2.33 & 312.9 & 0.82 & 0.59 & 10 & 1.0B & 0.60 & 8.21G \\
                           & Test-time Training \cite{ttt1} & 2.15 & 317.3 & 0.8  & 0.64 & 10 & 1.0B & 75.84 (+12,540\%) & 14.41G \\
\multirow{-3}{*}{VAR d-24 \cite{var}} & Composer           & \textbf{2.08} & \textbf{319.5} & \textbf{0.83} & \textbf{0.64} & \textbf{10} & \textbf{1.15B} & \textbf{0.63 (+5\%}) & \textbf{8.51G} \\ \midrule
                           & Standard           & 1.97 & 323.1 & 0.82 & 0.59 & 10 & 2.0B & 1 & 16.57G \\
                           & Test-time Training \cite{ttt1}& 1.85 & 327.7 & 0.81 & 0.62 & 10 & 2.0B & 112.37 (+11,137\%) & 28.41G \\
\multirow{-3}{*}{VAR d-30 \cite{var}} & Composer           & \textbf{1.79} & \textbf{330.4} & \textbf{0.83} & \textbf{0.63} & \textbf{10} & \textbf{2.2B} & \textbf{1.07 (+7\%)} & \textbf{16.97G} \\ \midrule
                           & Standard           & 5.02 & 167.2 & 0.75 & 0.57 & 250 & 458M & 31 & 3.4G \\
                           & Test-time Training \cite{ttt1}& 4.55 & 185.0 & 0.74 & 0.6  & 250 & 458M & 75.21 (+142\%) & 6.8G \\
\multirow{-3}{*}{DiT-L/2 \cite{dit}}  & Composer           & \textbf{4.41} & \textbf{192.4} & \textbf{0.76} & \textbf{0.59} & \textbf{250} & \textbf{560M} & \textbf{31.02 (+0.06\%)} & \textbf{3.5G} \\ \midrule
                           & Standard           & 2.27 & 278.2 & 0.83 & 0.57 & 250 & 675M & 45 & 6.1G \\
                           & Test-time Training \cite{ttt1}& 2.12 & 280.0 & 0.82 & 0.60 & 250 & 675M & 117.24 (+160\%) & 12.1G \\
\multirow{-3}{*}{DiT-XL/2 \cite{dit}} & Composer           & \textbf{2.06} & \textbf{285.6} & \textbf{0.84} & \textbf{0.58} & \textbf{250} & \textbf{825M} & \textbf{45.03 (+0.07\%)} & \textbf{6.4G} \\ \bottomrule
\end{tabular}
}
\label{tab:class-condition}
\end{table*}

\begin{table}[t]
\centering
\caption{Comparison in ImageNet $512 \times 512$.}
\label{tab:class-condition-512}
\begin{adjustbox}{width=\linewidth}
\begin{tabular}{l l c c r r}
\toprule
Backbone & Method & FID & IS & Time & Memory \\
\midrule
\multirow{3}{*}{VAR d-36 \cite{var}} 
 & Standard & 2.63 & 303.2 & 1.00 & 21.56G \\
 & Test-time Training & 2.54 & 305.4 & 112.37 (+11,100\%) & 32.45G \\
 & Composer & \textbf{2.51} & \textbf{305.7} & \textbf{1.07 (+7\%)} & \textbf{21.96G} \\
\midrule
\multirow{3}{*}{DiT-XL/2 \cite{dit}} 
 & Standard & 3.04 & 240.8 & 81.00 & 8.3G \\
 & Test-time Training & 2.87 & 245.2 & 152.24 (+87\%) & 14.3G \\
 & Composer & \textbf{2.81} & \textbf{248.1 }& \textbf{81.03 (+0.04\%)} & \textbf{8.5G }\\
\bottomrule
\end{tabular}
\end{adjustbox}
\end{table}

\noindent \textbf{Datasets:} We use the ImageNet dataset for class-conditional image generation at resolutions of $256\times256$ and $512\times512$. All models are evaluated on the standard ImageNet validation set.  

\noindent \textbf{Backbone Models:} We consider various generative model architectures including VAR (d-16, d-24, d-30) and DiT (L/2, XL/2). For each backbone, we compare three approaches: (i) Standard, the baseline generative model; (ii) Test-time Training (TTT), where the model is fine-tuned on input-specific data at inference; and (iii) Composer, our proposed framework that dynamically composes instance-specific parameters.  

\noindent \textbf{Discussion:} Table~\ref{tab:class-condition} reports results on ImageNet $256\times256$, showing that Composer consistently outperforms both Standard and Test-time Training (TTT) across all model scales. For instance, on VAR d-16, Composer reduces FID from 3.55 (Standard) and 3.22 (TTT) to \textbf{3.15}, while improving IS and maintaining precision and recall. Similar trends hold for larger backbones (VAR d-24/d-30, DiT-L/2, DiT-XL/2), confirming Composer’s ability to produce more realistic and diverse generations.  

In addition, Composer incurs negligible computational cost. As shown in Table~\ref{tab:class-condition}, peak inference memory and runtime remain comparable to Standard, while TTT is significantly more expensive due to repeated fine-tuning. For example, on VAR d-30, Composer requires only \textbf{1.07s} and \textbf{16.97G}, compared to TTT’s 11.12s and 28.41G.  

Table~\ref{tab:class-condition-512} extends the comparison to $512\times512$ resolution and larger models (VAR d-36, DiT-XL/2). Composer again achieves the best trade-off between quality and efficiency, improving FID to \textbf{2.51} on VAR d-36 and \textbf{2.81} on DiT-XL/2, while keeping inference cost nearly identical to Standard.  

Overall, Composer delivers \textbf{consistent gains in generation quality} with \textbf{minimal overhead}, establishing it as a scalable and efficient framework for high-fidelity class-conditional image generation.

\noindent\textbf{Computational and Memory Analysis:} We evaluate the efficiency of Composer in terms of inference time and peak memory, comparing it to Standard and Test-time Training (TTT). As reported in Tables~\ref{tab:class-condition} and~\ref{tab:class-condition-512}, Composer introduces negligible overhead while consistently improving generation quality. For instance, on VAR d-16 ($256\times256$), Composer increases inference time from 0.40s to 0.42s (+0.2\%) and peak memory from 2.37G to 2.57G (+3.6\%), whereas TTT requires 40.52s (+10,030\%) and 4.58G (+93\%).

This efficiency is maintained across all backbones and resolutions. On larger models such as VAR d-30 and DiT-XL/2, Composer keeps inference time and memory nearly identical to Standard (e.g., VAR d-30: 1.07s, 16.97G vs. Standard 1.00s, 16.57G), while TTT incurs significant computational cost due to per-instance gradient updates. Similarly, at $512\times512$ resolution, Composer achieves improved FID (e.g., VAR d-36: 2.51) with only minor increases in time and memory, whereas TTT consumes over 100× more time and nearly 50\% more memory.

These results highlight that Composer’s instance-specific parameter composition is highly efficient: low-rank updates generated by a lightweight meta-generator enable adaptive inference without iterative optimization or large memory duplication. Consequently, Composer is a practical solution for high-resolution and large-scale generative models, combining quality improvements with scalability.

\subsection{Text-to-Image Generation}

\begin{table}[h!]
\centering
\caption{Comparison of text-to-image generation performance on Stable Diffusion 2.1 (SD2.1) using FID-30K and CLIP-30K metrics.}
\label{tab:t2i}
\begin{adjustbox}{width=\linewidth}
\begin{tabular}{l l c c c c}
\toprule
Backbone & Method & FID-30K & CLIP-30K & Time & Memory \\
\midrule
\multirow{3}{*}{SD2.1 \cite{ldm}} 
 & Standard & 13.45 & 0.30 & 1.00 & 8G \\
 & Test-time Training & 13.15 & 0.31 & 41.52 (+4,000\%) & 55G \\
 & Composer & \textbf{13.07} & \textbf{0.32} & 1.02 (+2\%) & 8.4G \\
\bottomrule
\end{tabular}
\end{adjustbox}
\end{table}

\noindent \textbf{Datasets:} We evaluate our method on the MS-COCO 2014 dataset, following the standard 30K prompt evaluation protocol used in prior works. The metrics include FID-30K and CLIP-30K, which assess image fidelity and text–image semantic alignment respectively.  

\noindent \textbf{Discussion:} Table~\ref{tab:t2i} summarizes the quantitative results on SD2.1. Composer consistently improves generation quality over both Standard and TTT. Specifically, it reduces FID-30K from 13.45 (Standard) and 13.15 (TTT) to \textbf{13.07}, while increasing CLIP-30K from 0.30 to \textbf{0.32}, indicating better text–image correspondence and overall fidelity.  

In terms of efficiency, Composer remains nearly as lightweight as the Standard model, requiring only \textbf{1.02s} inference time and \textbf{8.4G} peak memory—far lower than TTT, which demands 41.52s and 55G.  

Overall, these results demonstrate that Composer generalizes effectively to text-to-image generation, delivering \textbf{higher semantic alignment and visual quality} with \textbf{minimal computational overhead}, making it a practical framework for adaptive, efficient diffusion-based image synthesis.

\subsection{Composer for Quantized Diffusion Models}

\begin{table}[t]
\centering
\caption{Quantized diffusion model performance with and without Composer. W/A denotes weight/activation bitwidth. Metrics include IS, FID, sFID, and Precision.}
\label{tab:quantized-composer}
\begin{adjustbox}{max width=\linewidth}
\begin{tabular}{clcccc}
\toprule
Bitwidth (W/A) & Baseline & IS & FID & sFID & Precision \\
\midrule
32/32 & Full Precision & 364.73 & 11.28 & 7.7 & 93.66 \\
\midrule
4/8 & Q-Diffusion \cite{q-diffusion} & 336.8 & 9.29 & 9.29 & 91.06 \\
    & + Composer & 347.4 & 8.95 & 8.92 & 92.35 \\\midrule
4/8 & CTEC \cite{ctec} & 355.62 & 8.52 & 7.31 & 93.81 \\
    & + Composer & 359.2 & 8.25 & 7.11 & 94.15 \\
\midrule
2/8 & Q-Diffusion \cite{q-diffusion} & 49.08 & 43.36 & 17.15 & 43.18 \\
    & + Composer & 78.21 & 35.26 & 14.5 & 55.2 \\\midrule
2/8 & CTEC \cite{ctec} & 176.37 & 7.43 & 7.98 & 80.2 \\
    & + Composer & 191.74 & 7.11 & 7.45 & 82.5 \\
\bottomrule
\end{tabular}
\end{adjustbox}
\end{table}

\noindent \textbf{Datasets and Setting:} We evaluate Composer on COCO 2014 for diffusion models under post-training quantization. Bitwidths for weights and activations (W/A) vary across experiments, including 32/32 (full precision), 4/8, and 2/8. We compare standard quantized backbones (Q-Diffusion and CTEC) with their Composer-augmented counterparts. Metrics include Inception Score (IS), FID, sFID, and Precision (\%).

\noindent \textbf{Discussion:} Table~\ref{tab:quantized-composer} shows that Composer consistently improves generative quality and stability under quantization. For 4/8 bit models, Composer increases IS and reduces FID and sFID for both Q-Diffusion and CTEC, while slightly improving Precision. For extreme 2/8 quantization, Composer provides dramatic improvements: Q-Diffusion + Composer increases IS from 49.08 to 78.21 and Precision from 43.18\% to 55.2\%, while reducing FID and sFID substantially. CTEC + Composer shows similar gains, demonstrating that Composer effectively compensates for quantization-induced degradation.

\noindent Overall, these results highlight that Composer enables high-fidelity, quantization-aware generation, restoring quality while maintaining computational efficiency, and is effective across different backbone architectures and extreme low-precision settings.

\subsection{Composer for Test-Time Scaling}

\begin{table}[t]
\centering
\caption{Comparison of SD2.1 variants on COCO 2014 (30K samples). Metrics include FID, CLIP similarity, inference time (s), and peak GPU memory.}
\label{tab:sd2-composer}
\begin{adjustbox}{max width=0.9\linewidth}
\begin{tabular}{lcccc}
\toprule
Method & FID-30K & CLIP-30K & Time & Memory \\
\midrule
SD2.1 \cite{ldm} & 13.45 & 0.30 & 1 & 8G \\
+ Composer & 13.07 & 0.32 & 1.02 (+2\%) & 8.4G (+5\%) \\
+ ORM \cite{image-tts} & 13.15 & 0.33 & 20 & 8G \\
+ ORM + Composer & 12.87 & 0.34 & 20.02 (+0.1\%) & 8.4G (+5\%) \\
+ PARM \cite{image-tts} & 13.07 & 0.33 & 20 & 8G \\
+ PARM + Composer & 12.82 & 0.34 & 20.02 (+0.1\%) & 8.4G (+5\%) \\
\bottomrule
\end{tabular}
\end{adjustbox}
\end{table}

Table~\ref{tab:sd2-composer} shows that Composer consistently improves generation quality across all settings. For the baseline SD2.1, Composer reduces FID from 13.45 to \textbf{13.07} and increases CLIP from 0.30 to 0.32, with minimal overhead (+2\% runtime, +5\% memory). When applied on top of ORM \cite{image-tts} and PARM \cite{image-tts} backbones, Composer further improves FID and CLIP: SD2.1 + ORM + Composer achieves FID \textbf{12.87} and CLIP 0.34, while SD2.1 + PARM + Composer reaches FID \textbf{12.82} and CLIP 0.34, with negligible increase in inference cost. Overall, these results demonstrate that Composer reliably enhances generation quality and semantic alignment for SD2.1, while maintaining efficiency across different model augmentations.

\subsection{Ablation Studies}

\begin{figure}[t]
\begin{center} 
\includegraphics[width=0.85\linewidth]{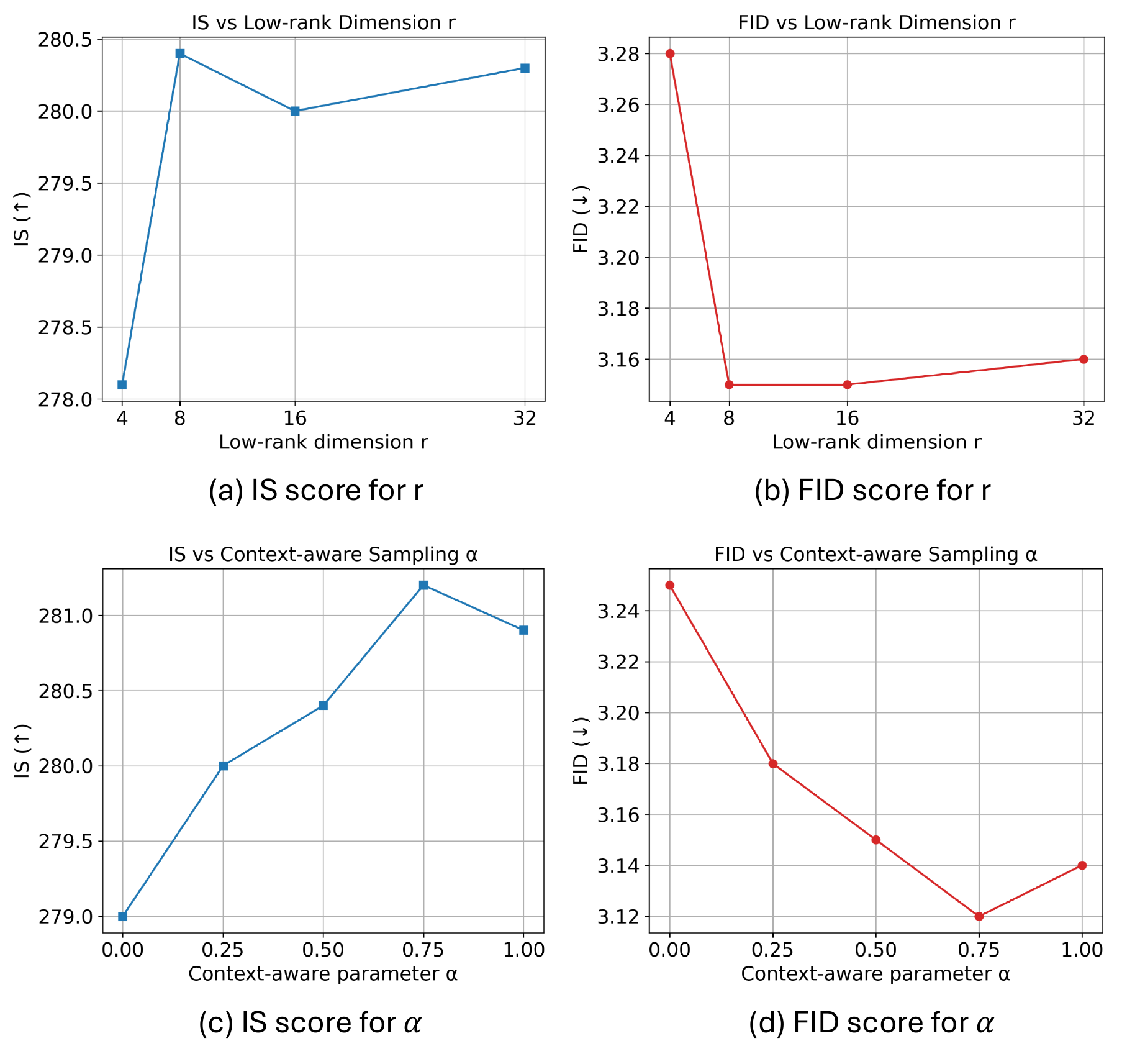}
\end{center}
\caption{Impact of low-rank dimension $r$ and context-aware sampling parameter $\alpha$ on ImageNet $256\times256$ class-conditional image generation. (a) Inception Score (IS) vs. $r$, (b) FID vs. $r$, (c) IS vs. $\alpha$, (d) FID vs. $\alpha$. Higher IS and lower FID indicate better generation quality.}
\label{fig:r-a}
\end{figure}

\noindent\textbf{Low-rank Dimension $r$:} We evaluate the effect of the low-rank dimension $r$ on class-conditional image generation using ImageNet $256\times256$. Figures~\ref{fig:r-a}(a-b) illustrate the IS and FID, respectively, for different values of $r$ ($4, 8, 16, 32$). As shown, increasing $r$ generally improves IS while slightly reducing FID, indicating that a larger low-rank dimension enhances both fidelity and diversity of generated images.  

\noindent\textbf{Context-aware Sampling Parameter $\alpha$:} We further analyze the impact of the context-aware sampling parameter $\alpha$ on generation quality. Figures~\ref{fig:r-a}(c-d) show IS and FID for $\alpha$ values from $0.0$ to $1.0$. Moderate values of $\alpha$ (around $0.5$-$0.75$) achieve the best balance between IS and FID, suggesting that context-aware sampling improves both realism and diversity without overfitting to context-specific prompts.  

\noindent\textbf{Standard vs Global-Local Attention.} We evaluate the impact of different attention mechanisms on class-conditional image generation for VAR d-16 and VAR d-24. The attention variants include Standard Attention, Local Attention, and Global-Local Attention (first block global, rest local). Table~\ref{tab:attention-comparison} reports FID and IS metrics for each setup. From Table~\ref{tab:attention-comparison}, Global-Local Attention achieves the lowest FID and highest IS on both VAR d-16 and VAR d-24, demonstrating that combining global and local context improves generative quality compared to purely local or standard attention.  

\begin{table}[ht]
\centering
\caption{FID and IS comparison of attention mechanisms on VAR d-16 and VAR d-24 for ImageNet $256\times256$.}
\label{tab:attention-comparison}
\adjustbox{max width=0.8\linewidth}{
\begin{tabular}{lcc|cc}
\toprule
\multirow{2}{*}{Attention} & \multicolumn{2}{c|}{VAR d-16} & \multicolumn{2}{c}{VAR d-24} \\
                            & FID $\downarrow$ & IS $\uparrow$ & FID $\downarrow$ & IS $\uparrow$ \\
\midrule
Standard Attention          & 3.55             & 274.4          & 2.33            & 312.9 \\
Global-Local Attention      & \textbf{3.15}    & \textbf{280.4} & \textbf{2.08}    & \textbf{319.5} \\
\bottomrule
\end{tabular}
}
\end{table}

\noindent\textbf{Comparing Our Transformer-Based Generator vs Other Architectures.} We evaluate the impact of different backbone architectures on class-conditional image generation for VAR d-16 and VAR d-24. Architectures include CNN-based, MLP-based, and Transformer-based generators. Table~\ref{tab:arch-comparison} reports FID and IS metrics for each setup. From Table~\ref{tab:arch-comparison}, the Transformer-based generator clearly outperforms CNN- and MLP-based architectures on both VAR d-16 and VAR d-24, achieving the lowest FID and highest IS. This indicates that Transformers are more effective at modeling complex image distributions, producing more realistic and diverse class-conditional images for ImageNet $256\times256$.

\begin{table}[ht]
\centering
\caption{FID and IS comparison of different generator architectures on VAR d-16 and VAR d-24 for ImageNet $256\times256$.}
\label{tab:arch-comparison}
\adjustbox{max width=0.8\linewidth}{
\begin{tabular}{lcc|cc}
\toprule
\multirow{2}{*}{Architecture} & \multicolumn{2}{c|}{VAR d-16} & \multicolumn{2}{c}{VAR d-24} \\
                              & FID $\downarrow$ & IS $\uparrow$ & FID $\downarrow$ & IS $\uparrow$ \\
\midrule
CNN-based Generator           & 3.35             & 276.1          & 2.18            & 315.2 \\
MLP-based Generator           & 3.32             & 275.4          & 2.21             & 315.0 \\
Transformer-based Generator   & \textbf{3.15}    & \textbf{280.4} & \textbf{2.08}    & \textbf{319.5} \\
\bottomrule
\end{tabular}
}
\end{table}

\noindent\textbf{Comparing Different Training and Sampling Pipelines.} Tables~\ref{tab:arch-comparison} and \ref{tab:t2i} report the performance of different generative frameworks in class-conditional and text-to-image settings, respectively. For ImageNet $256\times256$, Table~\ref{tab:arch-comparison} shows that the Transformer-based generator consistently outperforms CNN- and MLP-based architectures for VAR d-16 and VAR d-24, achieving the lowest FID and highest IS. This indicates that Transformers better model complex image distributions, producing more realistic and diverse class-conditional generations.  

In the text-to-image setting on COCO 2014 (Table~\ref{tab:t2i}), the Context-Aware Pipeline using Composer achieves the best FID-30K and CLIP-30K scores, outperforming both Vanilla Training and Similar-Prompt Sampling. This shows that Composer improves generation quality and text–image alignment while remaining efficient.

\begin{table}[ht]
\centering
\caption{FID and IS comparison of different training/sampling pipelines on VAR d-16 and VAR d-24 for ImageNet $256\times256$.}
\label{tab:pipeline-comparison}
\adjustbox{max width=0.75\linewidth}{
\begin{tabular}{lcc|cc}
\toprule
\multirow{2}{*}{Pipeline} & \multicolumn{2}{c|}{VAR d-16} & \multicolumn{2}{c}{VAR d-24} \\
                           & FID $\downarrow$ & IS $\uparrow$ & FID $\downarrow$ & IS $\uparrow$ \\
\midrule
Vanilla Training            & 3.55             & 274.4          & 2.33             & 312.9 \\
Full-Class & 3.28 & 277.6 & 2.18 & 316.5 \\
Context-Aware Pipeline       & \textbf{3.15}    & \textbf{280.4} & \textbf{2.08}    & \textbf{319.5} \\
\bottomrule
\end{tabular}
}
\end{table}

\begin{table}[h!]
\centering
\caption{Comparison of text-to-image generation pipelines on SD2.1 using COCO 2014.}
\label{tab:t2i}
\begin{adjustbox}{width=0.75\linewidth}
\begin{tabular}{l c c}
\toprule
Method & FID-30K $\downarrow$ & CLIP-30K $\uparrow$ \\
\midrule
Vanilla Training & 13.45 & 0.30 \\
Similar-Prompt Sampling & 13.15 & 0.31 \\
Context-Aware Pipeline & \textbf{13.07} & \textbf{0.32} \\
\bottomrule
\end{tabular}
\end{adjustbox}
\end{table}

Additional visualization, ablation studies and results can be found at \textbf{Supplementary Material}.

\section{Conclusion}

We introduced \textbf{Composer}, a plug-in framework that turns static diffusion and visual auto-regressive models into adaptive generators by composing lightweight, low-rank parameter updates per input. Composer sidesteps expensive test-time training, providing fine-grained, instance-specific adaptation on top of frozen pretrained backbones with negligible computational and memory overhead. Experiments on various image generation task including quantization and test-time scalings show consistent performance gains. By enabling input-aware parameter composition for off-the-shelf generators, Composer moves generative modeling toward genuinely adaptive, context-sensitive behavior.

\noindent\textbf{Limitation and Future work.} Despite these gains, Composer introduces a small amount of additional inference time and memory usage. Its efficiency depends on the number of generation steps: for models with many steps, such as DiT with 250 iterations, the overhead remains below 0.1\%, whereas for models with very few steps, such as one-step diffusion, the relative overhead becomes more noticeable. In future work, we aim to extend this paradigm to broader vision tasks, including transfer learning for generative modeling, where only Composer is fine-tuned while the backbone remains frozen. 

\section*{Acknowledgements}
Trung Le, Mehrtash Harandi, and Dinh Phung were supported by the ARC Discovery Project grant DP250100262. Trung Le and Mehrtash Harandi were also supported by the Air Force Office of Scientific Research under award number FA9550-23-S-0001.

{
    \small
    \bibliographystyle{ieeenat_fullname}
    \bibliography{main}
}

\

\clearpage
\setcounter{page}{1}
\maketitlesupplementary

\begin{strip}\centering
\includegraphics[width=\textwidth]{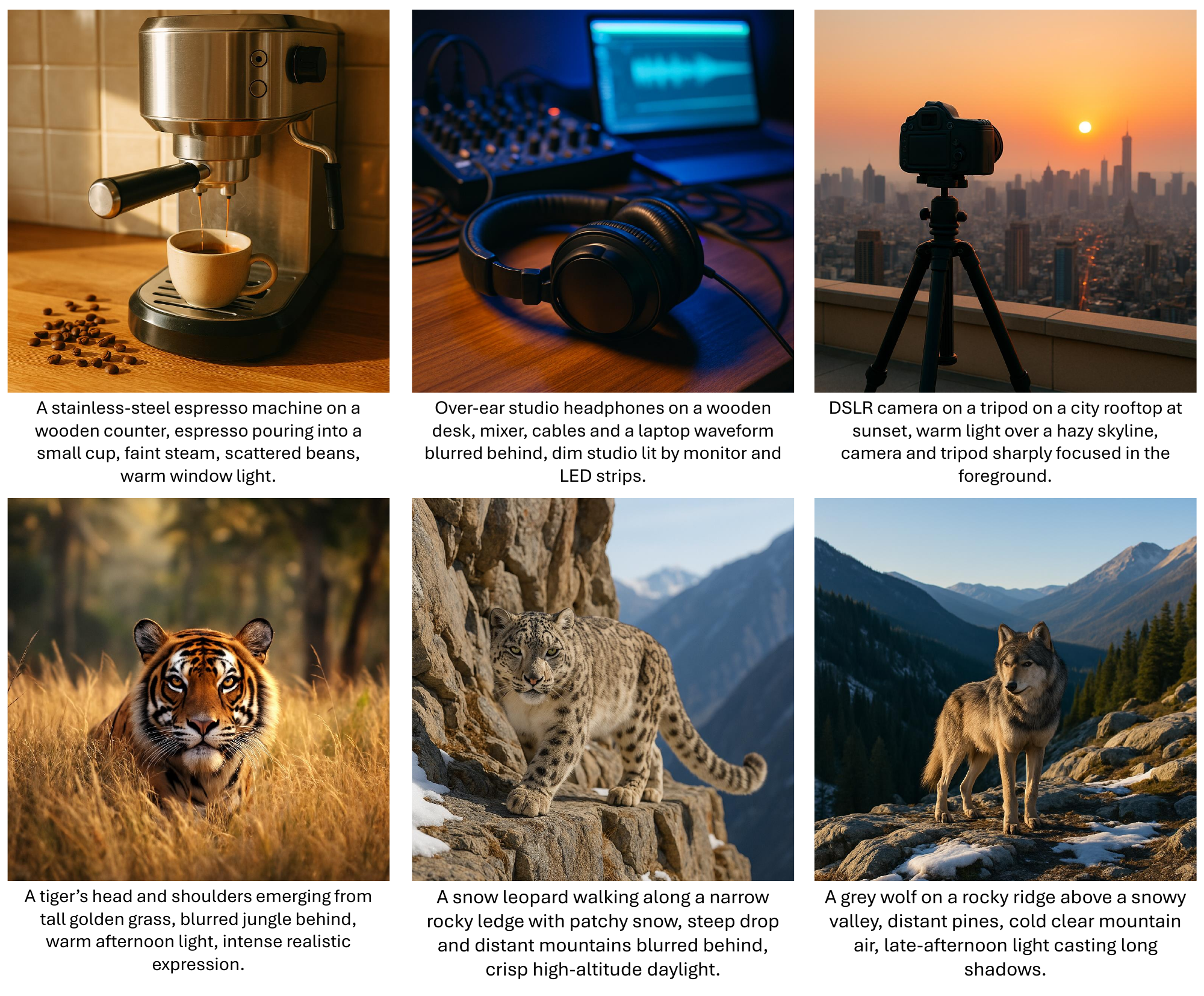}
\captionof{figure}{Qualitative comparison on additional text-to-image examples. For each prompt, we show images produced by baseline methods and by our approach.
\label{fig:visual2}}
\end{strip}

\section{Visualization}
To better understand how Composer affects the generative process beyond aggregate metrics, we provide additional qualitative examples in Figure~\ref{fig:visual2}. For each text prompt, we juxtapose images produced by competitive baselines with those generated by our approach. Across a wide range of prompts, including both simple single-object descriptions and more challenging compositional or attribute-heavy queries, our method consistently produces samples that better match the specified content.

Visually, Composer tends to preserve fine-grained object structure and attributes (e.g., pose, color, accessories) while also producing more coherent and contextually appropriate backgrounds. In contrast, baseline models often miss attributes, distort object shapes, or introduce clutter and artifacts in the scene. The improvements are especially noticeable for prompts that require binding multiple constraints (such as specific styles, actions, or environments), where our instance-specific parameter composition yields sharper details and more faithful semantic alignment. These qualitative results complement our quantitative findings and highlight that Composer enhances both realism and prompt adherence in text-to-image generation.

\section{Additional Ablation Studies.}

\subsection{Component Analysis}

We provided the ablation to isolate each component (see Table \ref{tab:composer_ablation_component_1}). Both the Context-Aware Pipeline and Global-Local Attention individually improve FID/IS, and combining them (full Composer) yields the best performance, indicating complementary gains.

\begin{table}[ht]
\centering
\caption{Ablation isolating the two improvements in Composer: (i) Context-Aware Pipeline and (ii) Global-Local Attention. All settings use the same VAR backbone and training budget; we add one component at a time, then combine them.}
\label{tab:composer_ablation_component_1}
\adjustbox{max width=\linewidth}{
\begin{tabular}{lcc|cc}
\toprule
\multirow{2}{*}{Method} 
& \multicolumn{2}{c|}{VAR d-16} 
& \multicolumn{2}{c}{VAR d-24} \\
& FID $\downarrow$ & IS $\uparrow$ & FID $\downarrow$ & IS $\uparrow$ \\
\midrule
Composer & {\color{red}\textbf{3.15}}
& {\color{red}\textbf{280.4}}
& {\color{red}\textbf{2.08}}
& {\color{red}\textbf{319.5}} \\

Composer -- Context-Aware Pipeline 
& 3.28 & 277.6 & 2.18 & 316.5 \\

Composer -- Global-Local Attention 
& 3.26 & 276.2 & 2.16 & 314.9 \\

Composer -- Context-Aware Pipeline -- Global-Local Attention

& 3.55 & 274.4 & 2.33 & 312.9 \\

\bottomrule
\end{tabular}
}
\end{table}

\subsection{Token initialization:} We provide an ablation to quantify the benefit of different type of token initialization in Table \ref{tab:token_init_ablation}.

\begin{table}[h]
\centering
\setlength{\tabcolsep}{3.2pt}
\caption{\textbf{Ablation on token initialization.} Comparing our weight-projected initialization/anchoring vs. using constant learnable tokens (trained as embeddings). Lower FID / higher IS is better.}
\label{tab:token_init_ablation}
\begin{adjustbox}{width=0.8\linewidth}
\begin{tabular}{l|cc|cc}
\toprule
\multirow{2}{*}{Method} &
\multicolumn{2}{c|}{VAR d-16} &
\multicolumn{2}{c}{VAR d-24} \\
\cmidrule(lr){2-3}\cmidrule(lr){4-5}
& FID$\downarrow$ & IS$\uparrow$ & FID$\downarrow$ & IS$\uparrow$ \\
\midrule
Constant learnable tokens (embeddings) 
& 3.17 & 279.8 & 2.07 & 318.7 \\
\midrule
Ours (projected init/anchored to frozen weights) 
& {\color{red}\textbf{3.15}} & {\color{red}\textbf{280.4}} & {\color{red}\textbf{2.08}} & {\color{red}\textbf{319.5}} \\

\bottomrule
\end{tabular}
\end{adjustbox}
\end{table}

\subsection{Where to apply Composer: which layers / which weights?}

We next study where Composer should be applied inside the self-attention blocks by ablating which projection matrices are modified. Concretely, we inject instance-specific low-rank updates into different subsets of the attention projections $\{W_Q, W_K, W_V, W_O\}$ and report results on ImageNet $256\times256$ for VAR d-16 and d-24 in Table~\ref{tab:layers-ablation}. Adapting only a single projection ($W_Q$, $W_K$, or $W_V$) already yields competitive performance, with $W_V$ slightly outperforming $W_Q$ and $W_K$, suggesting that modulating the value content is particularly important. Combining projections consistently improves results: variants that include $W_V$ such as $W_Q{+}W_V$ and $W_K{+}W_V$ are stronger than $W_Q{+}W_K$, and our default choice $W_Q{+}W_V$ achieves the best or near-best FID/IS across both backbones (e.g., $3.15$ FID / $280.4$ IS for VAR d-16 and $2.08$ FID / $319.5$ IS for VAR d-24). Extending Composer to all projections ($W_Q{+}W_K{+}W_V{+}W_O$) brings only marginal changes relative to $W_Q{+}W_V$ while increasing the number of adapted parameters and memory footprint. We therefore adopt adapting $W_Q$ and $W_V$ as the default configuration in all main experiments, as it provides the best accuracy–efficiency trade-off.

\begin{table}[ht]
\centering
\caption{Ablation on modified attention layers for VAR d-16 and VAR d-24 on ImageNet $256\times256$.}
\label{tab:layers-ablation}
\adjustbox{max width=\linewidth}{
\begin{tabular}{lcccc}
\toprule
\multirow{2}{*}{Layers} & \multicolumn{2}{c|}{VAR d-16} & \multicolumn{2}{c}{VAR d-24} \\
                            & FID $\downarrow$ & IS $\uparrow$ & FID $\downarrow$ & IS $\uparrow$ \\
\midrule
$W_Q$                     & 3.29 & 274.4 & 2.18 & 312.9 \\
$W_V$                     & 3.28 & 277.6 & 2.18 & 316.5 \\
$W_K$                     & 3.32 & 273.8 & 2.21 & 311.2 \\
$W_Q{+}W_K$               & 3.23 & 276.1 & 2.13 & 315.1 \\
$W_K{+}W_V$               & 3.21 & 278.9 & 2.13 & 317.8 \\
\textbf{$W_Q{+}W_V$ (Our choice)} & \textbf{3.15} & \textbf{280.4} & \textbf{2.08} & \textbf{319.5} \\
$W_Q{+}W_K{+}W_V$         & 3.16 & 279.9 & 2.09 & 318.7 \\
$W_Q{+}W_K{+}W_V{+}W_O$   & 3.15 & 280.7 & 2.08 & 319.4 \\
\bottomrule
\end{tabular}
}
\end{table}

\subsection{How large should the Composer generator's $d_{\text{model}}$ be?}

We further investigate the capacity of the Composer meta-generator by ablating its transformer width $d_{\text{model}}$. We vary $d_{\text{model}} \in \{256, 512, 1024, 2048, 4096\}$ while keeping the backbone, low-rank dimension $r$, and all training settings fixed, and report results on ImageNet $256\times256$ for VAR d-16 and VAR d-24 in Table~\ref{tab:width-ablation}. As we increase $d_{\text{model}}$ from 256 to 1024, both backbones consistently benefit: FID steadily decreases and IS improves (e.g., from 3.21/276.1 to 3.15/280.4 for VAR d-16, and from 2.14/313.2 to 2.08/319.5 for VAR d-24). Beyond $d_{\text{model}}=1024$, further widening yields only marginal changes in FID and slightly fluctuating IS (e.g., VAR d-16 attains a slightly lower FID of 3.14 at 4096 but with lower IS than at 1024), while incurring additional parameters and training cost. Overall, $d_{\text{model}}=1024$ provides the best or near-best FID/IS across both backbones under a modest computational budget, and we therefore adopt it as the default Composer configuration in all main experiments.

\begin{table}[ht]
\centering
\caption{Ablation on Composer generator width $d_{\text{model}}$ for VAR d-16 and VAR d-24 on ImageNet $256\times256$.}
\label{tab:width-ablation}
\adjustbox{max width=\linewidth}{
\begin{tabular}{lcccc}
\toprule
\multirow{2}{*}{$d_{\text{model}}$} & \multicolumn{2}{c|}{VAR d-16} & \multicolumn{2}{c}{VAR d-24} \\
                                   & FID $\downarrow$ & IS $\uparrow$ & FID $\downarrow$ & IS $\uparrow$ \\
\midrule
256   & 3.21 & 276.1 & 2.14 & 313.2 \\
512   & 3.18 & 278.4 & 2.12 & 315.4 \\
\textbf{1024 (Our choice)} 
      & \textbf{3.15} & \textbf{280.4} & \textbf{2.08} & \textbf{319.5} \\
2048  & 3.15 & 279.9 & 2.09 & 318.9 \\
4096  & 3.14 & 279.4 & 2.09 & 318.4 \\
\bottomrule
\end{tabular}
}
\end{table}

\subsection{Composer for OOD image generation:}

We provide the additional experiment on personalized text-to-image generation in Table \ref{tab:pipeline-comparison}. Unlike TTT, which fine-tunes the model per subject, Composer directly composes subject-adapted weights at inference time without gradients, achieving stronger gains under the same test-time budget. This extension relies on single-subject training data, benefits from a larger Transformer backbone, and can further improve with optional image-caption descriptor conditioning.

\begin{table}[h]
\centering
\caption{FID and IS comparison of different training/sampling pipelines on VAR d-16 and VAR d-24 for ImageNet $256\times256$.}
\label{tab:pipeline-comparison}
\adjustbox{max width=\linewidth}{
\begin{tabular}{@{}llrrrrrrrr@{}}
\toprule
Category & Model 
& \multicolumn{1}{c}{FID $\downarrow$} 
& \multicolumn{1}{c}{IS $\uparrow$} 
& \multicolumn{1}{c}{Pre} 
& \multicolumn{1}{c}{Rec} 
& \multicolumn{1}{c}{Step} 
& \multicolumn{1}{c}{Parameter} 
& \multicolumn{1}{c}{Time} 
& \multicolumn{1}{c}{Memory} \\
\midrule
\multirow{2}{*}{VAR d-16 }
 & Test-time Training & 3.28 & 279.1 & 0.85 & 0.53 & 10 & 310M & 40.52 (+10,030\%) & 4.58G \\
 & Composer (ours) & {\color{red}\textbf{3.23}} & {\color{red}\textbf{281.4}} & {\color{red}\textbf{0.87}} & {\color{red}\textbf{0.56}} & {\color{red}\textbf{10}} & {\color{red}\textbf{412M}} & {\color{red}\textbf{0.42 (+5\%)}} & {\color{red}\textbf{2.57G}} \\
\bottomrule
\end{tabular}
}
\end{table}
\vspace{-12pt}

\end{document}